\documentclass[letterpaper]{article} 
\usepackage{aaai25}  
\usepackage{times}  
\usepackage{helvet}  
\usepackage{courier}  
\usepackage[hyphens]{url}  
\usepackage{graphicx} 
\usepackage{natbib}  
\usepackage{caption} 
\usepackage{algorithm}
\usepackage{algorithmic}

\usepackage{newfloat}
\usepackage{listings}
\usepackage{enumitem}
\usepackage{svg}
\usepackage{multirow}
\usepackage{array}
\usepackage{tabularx} 
\usepackage{booktabs} 
\usepackage{amsmath}
\usepackage{amssymb}
\usepackage[hang,flushmargin]{footmisc}
\DeclareCaptionStyle{ruled}{labelfont=normalfont,labelsep=colon,strut=off} 
\floatstyle{ruled}
\newfloat{listing}{tb}{lst}{}
\floatname{listing}{Listing}

\makeatletter
\long\def\@makefntext#1{%
  \parindent 0pt
  \leftskip 0pt
  \noindent
  \@makefnmark\ #1
}
\makeatother

\title{Multimodal Forecasting for Commodity Prices \\ Using Spectrogram-Based and Time Series Representations}
\author{
    Soyeon Park\textsuperscript{\rm 1}\thanks{Part of this work was carried out during Soyeon Park’s internship at Impactive AI.},
    Doohee Chung\textsuperscript{\rm 1,2},
    Charmgil Hong\textsuperscript{\rm 1,2}
}
\affiliations{
    \textsuperscript{\rm 1}Handong Global University,
    Pohang, Republic of Korea\\
    \textsuperscript{\rm 2}Impactive AI,
    Seoul, Republic of Korea\\
    \{soyeon.park, profchung, charmgil\}@handong.ac.kr\\
}

\begin{document}

\maketitle

\begin{abstract}
Forecasting multivariate time series remains challenging due to complex cross-variable dependencies and the presence of heterogeneous external influences.
This paper presents \textit{Spectrogram-Enhanced Multimodal Fusion} (SEMF), which combines spectral and temporal representations for more accurate and robust forecasting.
The target time series is transformed into Morlet wavelet spectrograms, from which a Vision Transformer encoder extracts localized, frequency-aware features.
In parallel, exogenous variables, such as financial indicators and macroeconomic signals, are encoded via a Transformer to capture temporal dependencies and multivariate dynamics.
A bidirectional cross-attention module integrates these modalities into a unified representation that preserves distinct signal characteristics while modeling cross-modal correlations.
Applied to multiple commodity price forecasting tasks, SEMF achieves consistent improvements over seven competitive baselines across multiple forecasting horizons and evaluation metrics.
These results demonstrate the effectiveness of multimodal fusion and spectrogram-based encoding in capturing multi-scale patterns within complex financial time series.
\end{abstract}

\section{Introduction}
Time series prediction plays a fundamental role in organizational decision-making across business domains, including finance, energy, and manufacturing. 
In particular, accurate forecasts of commodity prices, such as those of gold, crude oil, nickel, and aluminum, directly influence strategic planning, risk management, procurement, and hedging decisions~\cite{SEZER2020106181}. 
These forecasts inform not only short-term trading actions but also longer-term operational and investment strategies that require consistency across multiple time horizons.
However, decision-makers often face substantial uncertainty because commodity markets reflect complex interactions among macroeconomic indicators, policy shifts, geopolitical events, and supply chain disruptions.
Consequently, commodity price forecasting represents a decision-critical task in which forecast errors lead to inventory misallocation, hedging inefficiency, and delayed operational responses.

Commodity price series exhibit nonlinear and non-stationary behavior that arises from diverse market mechanisms and external drivers.
This heterogeneity manifests through shifts in volatility regimes, variations in frequency composition, and differing sensitivities to exogenous factors, which complicate the alignment between forecasts and business decisions.
Moreover, organizations typically rely on forecasts across multiple planning horizons, which places additional demands on model robustness and consistency.
These characteristics make cross-commodity generalization and multi-horizon reliability central challenges in business-oriented forecasting systems.

Traditional deep learning models for time series forecasting, such as Long Short-Term Memory (LSTM) networks~\cite{hochreiter1997long}, model sequential dependencies through recurrent architectures and have been widely applied to financial data.
While these approaches capture short-term temporal patterns effectively, they exhibit notable performance degradation in long-horizon settings with multiple temporal scales and frequency-dependent patterns~\cite{SEZER2020106181}.
Commodity price data exhibit abrupt fluctuations alongside long- and short-term dynamics that no single representation can capture.
This limitation is particularly problematic in business contexts where forecasts must remain reliable across operational and strategic horizons.

Recent studies have explored the transformation of time series data into image-based representations to exploit advances in computer vision models~\cite{semenoglou2023image}.
While line plot images enable the extraction of structural patterns, they provide limited access to frequency-domain characteristics that are essential for analyzing non-stationary volatility.
Alternative approaches reformulate forecasting as an image reconstruction task, demonstrating that pre-trained visual autoencoders can act as generic forecasters~\cite{chen2024visionts}.
Although these methods show promise, they often lack sufficient capacity to capture multiscale temporal dynamics and contextual dependencies, such as macroeconomic signals or market-wide risk indicators that influence business decisions.
As a result, vision-based representations alone remain insufficient for complex financial forecasting tasks.

Morlet wavelet spectrograms~\cite{torrence1998practical} offer a principled time-frequency representation that preserves localized spectral information across multiple scales.
Compared to simple visual representations such as line plot images or generic image-based encodings discussed above, Morlet spectrograms provide a substantially richer description of non-stationary signals by retaining scale-dependent energy distributions and phase information.
The Morlet wavelet, formed by modulating a Gaussian window with a complex sinusoid, yields complex-valued coefficients that enable precise characterization of transient oscillatory behavior and abrupt spectral shifts.
Prior studies have demonstrated the effectiveness of Morlet spectrograms as inputs for forecasting models~\cite{zeng2023pixels}.
However, existing approaches primarily focus on univariate signals processed by a single Vision Transformer (ViT)~\cite{dosovitskiy2020image}, which limits their ability to model multivariate dependencies and incorporate external contextual information.

To address these limitations, we propose \textbf{Spectrogram-Enhanced Multimodal Fusion} (SEMF), a dual-path framework that integrates spectral and temporal representations. 
SEMF employs a ViT that encodes Morlet wavelet spectrograms in order to capture scale-specific market dynamics, while a Transformer-based encoder processes reversible instance normalization (RevIN)-transformed exogenous time series that represent macroeconomic and financial context~\cite{kim2021reversible}. 
A bidirectional cross-attention module aligns the two modalities within a shared representation space, which facilitates interaction between time-frequency structures and contextual signals.
This design supports forecasts that remain stable across multiple horizons, where spectral market dynamics and external contextual signals must be considered jointly.
By jointly modeling spectral patterns and multivariate dependencies, SEMF addresses representational gaps that limit prior approaches.

We evaluate SEMF on commodity price forecasting tasks that involve assets with diverse market characteristics and external influences.
The experimental design includes macro-financial variables that reflect signals considered in managerial decision-making.
Experimental results show that SEMF achieves consistent performance improvements over conventional time series models and state-of-the-art image-based approaches across multiple forecasting horizons.
The findings reveal the robustness and practical relevance of SEMF in financial environments with high volatility and complex multivariate dynamics.
Such robustness across forecasting horizons supports stable risk management and coherent long-term hedging.
It reduces horizon-induced decision variability, limits unexpected losses, and improves cost efficiency under uncertainty.

Our contributions are summarized as follows:
\begin{itemize}[leftmargin=*,itemsep=0pt, topsep=0pt, partopsep=0pt, parsep=0pt, itemindent=0pt]
    \item We propose Spectrogram-Enhanced Multimodal Fusion (SEMF), a forecasting architecture that integrates spectral and temporal representations to support robust decision-oriented forecasting.
    \item We employ Morlet wavelet spectrograms that capture nonlinear and non-stationary dynamics relevant to volatile commodity markets.
    \item We adopt a bidirectional cross-attention fusion mechanism that aligns spectral patterns with exogenous contextual signals within a unified representation.
    \item We implement a multi-horizon, multi-task learning framework for business and financial decision-making that supports coordinated short-term and long-term decisions under uncertainty.
\end{itemize}

\section{Related Work}
Time series forecasting has progressed from classical statistical models to modern learning-based approaches.
Statistical methods such as ARIMA and Prophet~\cite{taylor2018forecasting} perform reliably on series with strong seasonal or trend components; however, they generalize poorly in financial settings where nonlinear dynamics and heterogeneous exogenous factors interact.
These limitations restrict their ability to represent the complex and non-stationary nature of financial time series.

Deep learning models have, therefore, attracted significant attention.
Recurrent architectures, such as LSTM and GRU, capture sequential dependencies, while Transformer-based models leverage self-attention to capture long-range temporal correlations~\cite{nie2022time}.
Despite their expressive capacity, these approaches often exhibit instability when applied to financial data that involve high-frequency noise, regime shifts, and complex cross-variable interactions.
This observation suggests that a single temporal representation remains insufficient for high-complexity forecasting tasks. 

Recent studies have explored image-based formulations of time series forecasting by converting sequential data into visual representations.
ViT architectures have demonstrated improved abstraction of temporal structures compared to convolutional models~\cite{li2023time}.
Existing image-based encodings primarily focus on shape and structural patterns, whereas explicit time-frequency resolution remains insufficient, which constrains the modeling of non-stationary volatility and multi-scale dynamics in financial time series.
To address this limitation, recent work adopts time-frequency representations based on Morlet wavelet spectrograms.
Zeng et al.~\cite{zeng2023pixels} encode spectrogram-based features with numerical value intensities through convolutional networks and ViT, although the approach remains confined to univariate series without external contextual variables.

Motivated by these observations, our work proposes a dual-path multimodal framework that integrates time-frequency representations with raw multivariate time series.
The target commodity price series is encoded through a ViT operating on Morlet wavelet spectrograms, while exogenous variables are processed by a Transformer-based encoder to model multivariate temporal dynamics.
A cross-attention mechanism aligns these feature representations within a unified feature space, enabling joint spectral and temporal modeling for financial forecasting.

\begin{figure*}[t]
    \centering
    \includegraphics[width=\linewidth]{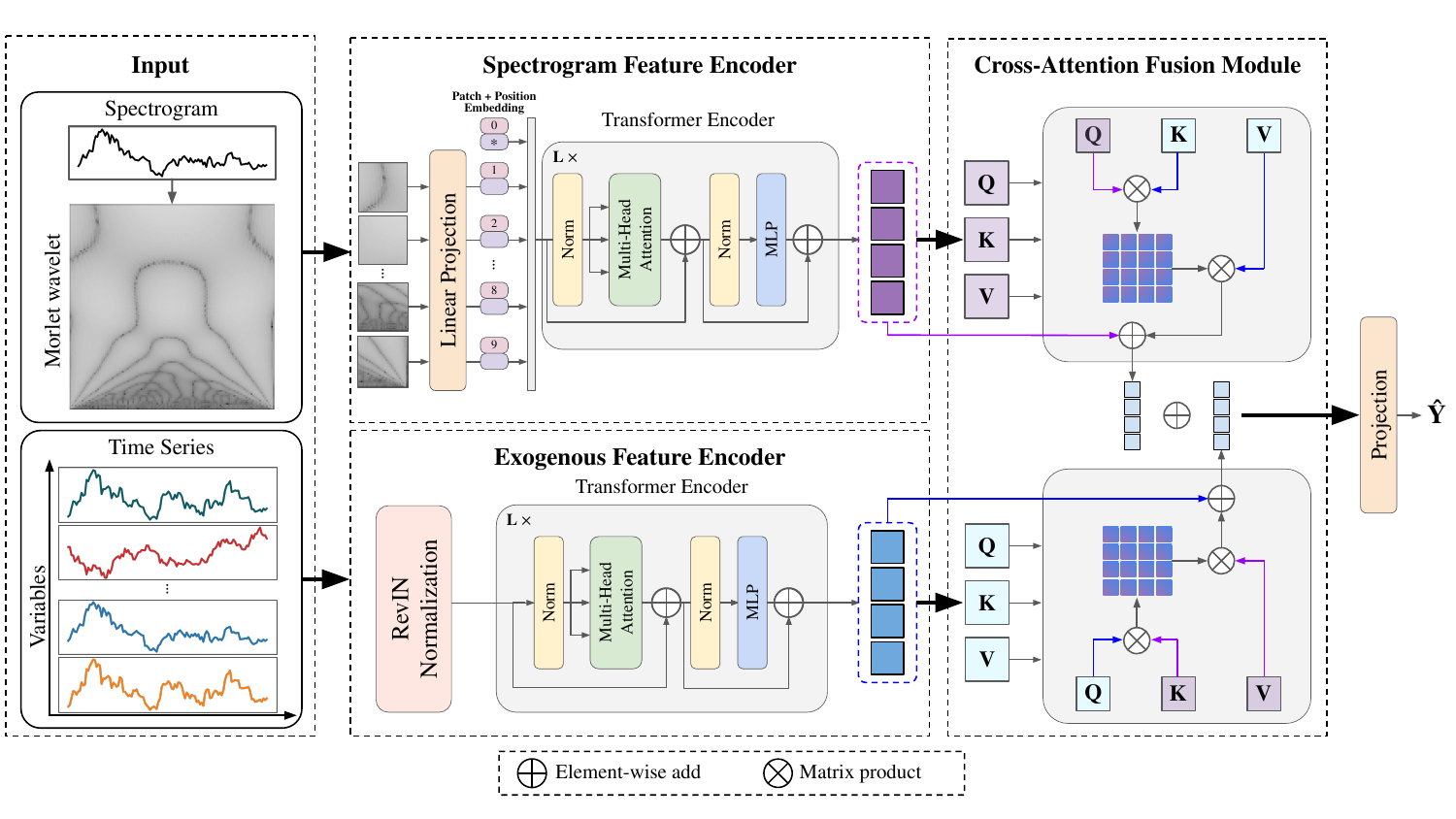}
    \caption{Overall framework of Spectrogram-Enhanced Multimodal Fusion (SEMF).}
    \label{fig:framework}
\end{figure*}

\section{Methodology}

\subsection{Problem Statement}
Let $\mathbf{X} = \{\mathbf{x}_1, \mathbf{x}_2, \ldots, \mathbf{x}_T\}$ denote a multivariate time series, where each observation $\mathbf{x}_t \in \mathbb{R}^D$ represents $D$ variables observed at time step $t$. 
The target series is defined as $y_t = x_t^{(1)}$, while the remaining variables form the exogenous time series $\mathbf{Z} = \{\mathbf{z}_1, \ldots, \mathbf{z}_T\}$ with $\mathbf{z}_t \in \mathbb{R}^{D-1}$.
Given historical observations $(\mathbf{y}_{1:T}, \mathbf{Z}_{1:T})$, the objective is to predict the target series at multiple future horizons,
$\mathbf{Y} = \{y_{T+1}, y_{T+3}, y_{T+7}, y_{T+14}, y_{T+21}, y_{T+35}\}$.
The model learns a mapping
$f_\theta : (\mathbf{y}_{1:T}, \mathbf{Z}_{1:T}) \rightarrow \hat{\mathbf{Y}}$,
where $\hat{\mathbf{Y}}$ denotes the predicted values and $\theta$ represents the model parameters. 
The goal of training is to minimize the Mean Squared Error (MSE) over all forecasting horizons, which serves as the primary optimization criterion throughout the experimental evaluation.

\subsection{Overall Framework}
The proposed Spectrogram-Enhanced Multimodal Fusion (SEMF) framework is designed to model localized frequency variations and long-range temporal dependencies in financial time series.
As illustrated in Figure~\ref{fig:framework}, SEMF consists of three main components: a Spectrogram Feature Encoder, an Exogenous Feature Encoder, and a Cross-Attention Fusion Module.
The framework transforms the target time series into a Morlet wavelet spectrogram, which is encoded by a ViT to capture time-frequency characteristics.
In parallel, the Exogenous Feature Encoder processes multivariate exogenous variables as sequential inputs to model temporal dynamics.
Finally, the Cross-Attention Fusion Module aligns these representations within a unified latent representation space, which enables joint spectral and temporal modeling for multi-horizon forecasting.
\subsection{Time Series Transformation}
This subsection describes the transformation procedures applied to the target and exogenous time series before feature encoding.
The target series is extracted from a fixed-length historical window and transformed into a time-frequency representation using the Morlet wavelet.
This transformation exposes localized fluctuations, multi-scale temporal patterns, and non-stationary dynamics that are difficult to discern in the raw time domain.
The resulting logarithmic amplitude spectrogram is standardized to a zero mean and unit variance before being input to the Spectrogram Feature Encoder.
This transformation enables effective modeling of both short- and long-term temporal dependencies.

The Morlet wavelet adopted in this study is defined as
$$\psi_0(\eta) = \pi^{-1/4} \, e^{i \omega_0 \eta} \, e^{-\eta^2 / 2}$$
where $\omega_0$ denotes the non-dimensional central frequency~\cite{torrence1998practical}.
The Morlet wavelet provides favorable theoretical and practical properties for analyzing non-stationary signals with oscillatory behavior.
In contrast to fixed-window time-frequency transforms such as the Short-Time Fourier Transform (STFT), wavelet-based representations offer adaptive resolution across scales, which enable fine temporal localization at high frequencies and improved frequency resolution at low frequencies.
This property aligns well with commodity price series, which exhibit abrupt short-term fluctuations together with longer-term cyclical dynamics.
Among commonly used wavelets, the Morlet wavelet achieves strong joint localization in time and frequency domains through its Gaussian envelope and complex exponential form.
Compared to real-valued wavelets such as Haar or Daubechies, the complex-valued Morlet wavelet preserves both amplitude and phase information, which facilitates the characterization of transient oscillatory behavior and abrupt spectral shifts.
These properties make the Morlet wavelet particularly suitable for representing localized frequency variations and regime transitions in non-stationary financial time series.

Exogenous variables are retained in their raw sequential form in order to preserve temporal structure and variable-specific distributions.
This design supports the Exogenous Feature Encoder in modeling cross-variable interactions, delayed effects, and short-term temporal patterns that complement the spectrogram-based representation.
For multi-horizon forecasting, target values corresponding to 1, 3, 7, 14, 21, and 35 days ahead are standardized individually.
As a result, the SEMF framework operates on input--output pairs that consist of normalized spectrograms, raw exogenous sequences, and standardized target values.
This unified preprocessing scheme enables the model to learn time-frequency representations and temporal dependencies within a consistent forecasting setting.

\subsection{Spectrogram Feature Encoder}
The Spectrogram Feature Encoder maps the spectrogram representation into a latent feature representation.
It employs a ViT-based architecture~\cite{dosovitskiy2020image}, in which the spectrogram is divided into patch tokens and encoded through multi-head self-attention.
This design allows the model to capture both local and global temporal-spectral interactions across the entire spectrogram.

Within the Transformer encoder, attention heads attend to complementary aspects of the spectrogram, allowing diverse structural patterns to be represented in the patch embeddings.
A dedicated CLS token aggregates global information, and the encoder produces a compact latent vector that summarizes the spectrogram content~\cite{vaswani2017attention}.
This representation serves as a frequency-aware summary that supports downstream multi-horizon forecasting.
The resulting latent vector is passed to the fusion module, where it is aligned with temporal representations derived from exogenous variables.
The attention-based structure promotes stable learning and facilitates effective modeling of interactions across different frequency patterns.

\subsection{Exogenous Feature Encoder}
The Exogenous Feature Encoder processes multivariate exogenous variables in their raw sequential representations and transforms them into temporal feature embeddings that capture complementary temporal dynamics.
These inputs are aligned with the same historical window as the target time series and reflect diverse external conditions. 
Each sequence is independently normalized using RevIN, which stabilizes training and mitigates distributional shifts while preserving variable-specific scale and behavior~\cite{kim2021reversible}.
This normalization preserves the temporal structure of each signal without introducing irreversible transformations.
As a result, the encoder retains informative dynamics that are essential for accurate multi-horizon forecasting.

The encoder employs a Transformer-based architecture to model the temporal dependencies within the exogenous signals~\cite{zhou2021informer, wu2021autoformer}. 
Through self-attention, it captures delayed effects, cross-variable interactions, and long-range temporal influences that may affect future movements of the target series.
This design enables the model to learn patterns that extend beyond local time steps and depend on interactions across variables.
The resulting embedding provides a context-rich summary of both short-term dynamics and global relationships, which are subsequently integrated with spectral representations in the fusion module.

\begin{table}[!t]
\centering
\begin{tabular}{l|l}
\hline
\textbf{Type} & \textbf{Variable} \\
\hline
Target &
Closing price of target commodity futures \\
\hline
\multirow{10}{*}{Exogenous}
& US 10-year Treasury yield \\
& US 2-year Treasury yield \\
& US 3-month Treasury bill yield \\
& US Dollar Index (DXY) \\
& USD to CNY exchange rate \\
& USD to JPY exchange rate \\
& USD to KRW exchange rate \\
& S\&P 500 index \\
& S\&P 500 VIX index (market volatility) \\
& LME Commodity Index \\
\hline
\end{tabular}
\caption{Input variables for commodity price forecasting.}
\label{tab:variables}
\end{table}

\begin{table*}[h]
\centering
\setlength{\tabcolsep}{8pt}
\renewcommand{\arraystretch}{1.1}
\begin{tabularx}{\linewidth}{>{\centering\arraybackslash}m{2.5cm}
                             >{\centering\arraybackslash}X
                             >{\centering\arraybackslash}X
                             >{\centering\arraybackslash}m{1.5cm}
                             >{\centering\arraybackslash}m{1.5cm}
                             >{\centering\arraybackslash}m{1.5cm}
                             >{\centering\arraybackslash}m{1.5cm}
                             }
\toprule
\textbf{Category} & \textbf{Model} & \textbf{Commodity} & \textbf{RMSE $\downarrow$} & \textbf{RMAE $\downarrow$} & \textbf{MAPE $\downarrow$} & \textbf{$\boldsymbol{\mathrm{R}}^2$ $\uparrow$} \\
\midrule
\multirow{12}{*}{Time series-based} 
    & \multirow{3}{*}{LSTM} 
                & Coal  & 18.86 & 0.1230 & 11.4718 & -0.1821\\
        &       & Gold  & 538.60 & 0.2220 & 0.2139 & -1.5574\\
        &       & Steel & 570.24 & 0.1491 & 15.6967 & -2.6951\\
        \cmidrule(lr){2-7}
    & \multirow{3}{*}{iTransformer} 
                & Coal  & 104.09 & 0.5663 & 55.7048 & -37.7016\\
        &       & Gold  & 112.52 & 0.0408 & 0.0407 &  \textbf{\textcolor{red}{0.9660}}\\
        &       & Steel & 183.48 & 0.0419 & 4.1703  & 0.5649\\
        \cmidrule(lr){2-7}
    & \multirow{3}{*}{TimesNet} 
                 & Coal  & 21.09 & 0.1052 & 10.4761 & -0.4346\\
        &       & Gold & \textbf{\textcolor{red}{104.23}} &  \underline{\textcolor{blue}{0.0378}} &  \underline{\textcolor{blue}{0.0369}} & \underline{\textcolor{blue}{0.9600}}\\
        &       & Steel & 199.92 &  \underline{\textcolor{blue}{0.0457}} & 4.6178 & 0.5463\\
        \cmidrule(lr){2-7}
    & \multirow{3}{*}{PatchTST} 
                & Coal  &  \underline{\textcolor{blue}{12.02}} &  \underline{\textcolor{blue}{0.0959}} & \textbf{\textcolor{red}{7.4217}} & \underline{\textcolor{blue}{0.5032}}\\
        &       & Gold  & 126.77 & 0.0568 & 0.0433 & 0.9545\\
        &       & Steel & \underline{\textcolor{blue}{177.35}} & 0.0527 &  \underline{\textcolor{blue}{4.1102}} & \underline{\textcolor{blue}{0.6494}}\\
\midrule
\multirow{6}{*}{Image-based} 
    & \multirow{3}{*}{VisionTS} 
                & Coal  & 70.77 & 0.5535 & 55.6418 & -15.6105\\
        &       & Gold  & 823.13 & 0.3279 & 0.3290 & -187.9199\\
        &       & Steel & 1522.21 & 0.4223 & 43.9314 & -23.9076 \\
        \cmidrule(lr){2-7}
    & \multirow{3}{*}{ViT-num-spec} 
                & Coal  & 45.10 & 0.3307 & 31.6826 & -5.5089\\
        &       & Gold  & 830.62 & 0.3525 & 0.3431 & -8.4687\\
        &       & Steel & 314.52 & 0.0835 & 8.4544 & -0.0619\\
\midrule
\multirow{3}{*}{Multimodal} 
    & \multirow{3}{*}{SEMF (Ours)} 
                & Coal  & \textbf{\textcolor{red}{11.85}} & \textbf{\textcolor{red}{0.0758}} &  \underline{\textcolor{blue}{7.4484}} &\textbf{\textcolor{red}{ 0.5500}}\\
        &       & Gold &  \underline{\textcolor{blue}{107.05}} & \textbf{\textcolor{red}{0.0361}} & \textbf{\textcolor{red}{0.0358}} & 0.8425\\
        &       & Steel & \textbf{\textcolor{red}{173.29}} & \textbf{\textcolor{red}{0.0409}} & \textbf{\textcolor{red}{4.1100}} & \textbf{\textcolor{red}{0.6734}} \\
\bottomrule
\end{tabularx}
\caption{Performance comparison across multiple forecasting horizons. Results are averaged over all horizons. \textbf{\textcolor{red}{Red}} and \underline{\textcolor{blue}{blue}} denote the best and second-best results, respectively.}
\label{tab:model_comparison}
\end{table*}

\subsection{Cross-Attention Fusion Module}
The Cross-Attention Fusion Module combines spectrogram-based and exogenous representations into a unified feature space for forecasting.
Although each encoder captures modality-specific characteristics, their independent outputs cannot represent the cross-modal dependencies required to model complex financial signals.
To address this limitation, the fusion module learns interactions between frequency-domain and time-domain features that are not accessible to either feature representation alone~\cite{tsai2019multimodal}.

The module applies a bidirectional cross-attention mechanism in which each modality alternately serves as the query while the other provides key–value pairs~\cite{vaswani2017attention}.
In one direction, a representative vector derived from the spectrogram sequence queries the exogenous feature sequence to relate frequency variations to external contextual signals~\cite{tsai2019multimodal, zhang2023crossformer}.
In the reverse direction, a summary vector from the exogenous sequence queries the spectrogram features to associate temporal dynamics with time-frequency patterns.
This bidirectional formulation enables symmetric information exchange and aligns complementary cues across modalities.

The outputs of the cross-attention layers are combined through residual connections and layer normalization to form a unified feature representation.
This design preserves modality-specific information, promotes training stability, and supports effective learning of cross-modal dependencies.
The resulting representation captures interdependencies across time and frequency and provides a compact summary that integrates complementary signals from both input pathways, supporting accurate multi-horizon forecasting.

\subsection{Multi-Horizon Forecasting}
The final fused representation generated by the Cross-Attention Fusion Module is passed through a shared two-layer MLP prediction head, consisting of LayerNorm, a linear projection, GELU activation, dropout, and a final linear layer, to produce six predictions corresponding to multiple horizons (1, 3, 7, 14, 21, and 35 days).
Instead of employing a single-step decoder or an autoregressive forecasting strategy, SEMF predicts all horizons simultaneously using a shared output layer. 
This design enables joint optimization across forecasting horizons and avoids error accumulation associated with recursive prediction.
The model is trained by minimizing the mean squared error, averaged across all forecasting horizons. 
By integrating frequency-sensitive representations with temporal interaction features, SEMF provides stable performance across both short- and long-term prediction horizons.


\section{Experiments}
We evaluate the effectiveness of the proposed SEMF framework through a series of empirical experiments.
First, we describe the experimental settings used for model training and testing.
We then compare SEMF against a set of strong baselines across multiple forecasting horizons and evaluation metrics.
Finally, we conduct ablation studies to examine the contributions of each architectural component and to validate the design choices behind the framework.

\subsection{Experimental Setting}
\subsubsection{Dataset}
We evaluate our model on a multivariate daily time series dataset constructed from commodity prices and macro-financial data from April 2013 to January 2026.
The dataset contains daily observations for multiple commodities and associated macroeconomic indicators, with all price-based variables aligned using daily closing prices to ensure consistency across assets.
Table~\ref{tab:variables} summarizes the input variables used for commodity price forecasting.
Missing values arising from data collection gaps and market-specific reporting differences are handled through an imputation procedure prior to model training.
This preprocessing step ensures complete and temporally aligned input sequences for all commodities considered in the study.

For each commodity, a fixed historical input window is constructed, and future prices at multiple horizons are predicted.
The dataset for each commodity is chronologically divided into training, validation, and test sets using a ratio of 0.65, 0.15, and 0.20, respectively.
This split results in a total of 3,185 samples per commodity, including 2,070 training samples, 478 validation samples, and 637 test samples.
Importantly, all commodities evaluated in this study share the same temporal coverage, preprocessing pipeline, and data split configuration.
This unified setup enables a fair and consistent comparison of forecasting performance across commodities with diverse market characteristics.

\subsubsection{Comparison Methods}
For SEMF, the input sequence length is set to 120, which is selected based on preliminary validation experiments to balance the representation of short- and long-term temporal patterns.
The Morlet wavelet uses a fixed scale of 128, resulting in a spectrogram output of size 128$\times$120 (scale$\times$sequence length).
The Spectrogram Feature Encoder adopts a ViT with a patch size of 8.
Detailed analyses of these selected hyperparameters and their effects on performance are presented in the Analysis.

The study compares the proposed SEMF model with seven representative baselines. 
The baselines include five time series models (LSTM~\cite{hochreiter1997long}, iTransformer~\cite{liu2023itransformer}, TimesNet~\cite{wu2022timesnet}, and PatchTST~\cite{nie2022time}) and two image-based forecasting models (VisionTS~\cite{chen2024visionts} and ViT-num-spec~\cite{zeng2023pixels}). 
These baselines represent diverse architectures and enable a systematic comparison of SEMF under different modeling assumptions.


\subsubsection{Evaluation Metrics}
We evaluate forecasting performance using four standard metrics: Root Mean Squared Error (RMSE), Relative Mean Absolute Error (RMAE), Mean Absolute Percentage Error (MAPE), and the coefficient of determination ($\mathrm{R}^2$).
RMSE captures absolute errors with higher sensitivity to large deviations, while RMAE and MAPE measure relative errors using scale-normalized and percentage-based schemes.
$\mathrm{R}^2$ quantifies the proportion of variance in the target series explained by the forecasts.
Metrics are computed independently for each forecasting horizon (1, 3, 7, 14, 21, and 35 days) and then averaged across horizons. 
Lower values indicate better performance for error-based metrics, whereas higher values are preferred for $\mathrm{R}^2$.

\subsection{Results}
Now we report the forecasting performance of the proposed SEMF framework, with an emphasis on robustness and stability across commodities and forecasting horizons, which are essential for decision-oriented forecasting.

Table~\ref{tab:model_comparison} summarizes forecasting performance across Coal, Gold, and Steel, with results averaged across all forecasting horizons.
Time series-based models demonstrate strong performance on individual commodities, although their accuracy varies substantially depending on asset-specific characteristics.
Image-based approaches consistently exhibit larger errors, which suggests limited suitability for modeling structured temporal dynamics in isolation.
Across all commodities and evaluation metrics, SEMF shows comparable error levels rather than sharp performance fluctuations.
Low relative error measures are observed for Gold, while similar accuracy levels are maintained for Coal and Steel under distinct volatility profiles.
The resulting performance pattern highlights robustness across heterogeneous market conditions rather than specialization to a single asset class.
In particular, for Gold, a discrepancy between RMSE and $\mathrm{R}^2$ is observed. 
This arises from their different sensitivity to target variance. 
While RMSE measures absolute error, $\mathrm{R}^2$ is normalized by horizon-specific variance.
As a result, inconsistent trends may appear when results are averaged across horizons.

\begin{table}[!t]
\centering
\renewcommand{\arraystretch}{1.1}
\begin{tabularx}{\linewidth}{
    >{\centering\arraybackslash}m{1.5cm}
    >{\centering\arraybackslash}m{0.5cm}
    >{\centering\arraybackslash}X
    >{\centering\arraybackslash}X
    >{\centering\arraybackslash}X
}
\hline
Model & H & RMSE & RMAE & MAPE \\
\hline
\multirow{7}{*}{\shortstack{TimesNet}} 
 & 1  & \textcolor{red}{\textbf{73.88}}  & \textcolor{red}{\textbf{0.0269}} & \textcolor{red}{\textbf{0.0262}} \\
 & 3  & \textcolor{red}{\textbf{89.43}}  & \textcolor{red}{\textbf{0.0325}} & \textcolor{red}{\textbf{0.0316}} \\
 & 7  & \textcolor{red}{\textbf{99.13}}  & 0.0363 & 0.0353 \\
 & 14 & 116.28 & 0.0426 & 0.0415 \\
 & 21 & 116.14 & 0.0418 & 0.0407 \\
 & 35 & 130.53 & 0.0469 & 0.0459 \\
 \hline
 & avg & \textcolor{red}{\textbf{104.23}} & 0.0378 & 0.0369 \\
\hline
\multirow{7}{*}{\shortstack{PatchTST}} 
 & 1  & 100.33 & 0.0455 & 0.0344 \\
 & 3  & 129.40 & 0.0586 & 0.0453 \\
 & 7  & 106.22 & 0.0479 & 0.0366 \\
 & 14 & 127.61 & 0.0572 & 0.0424 \\
 & 21 & 135.95 & 0.0606 & 0.0472 \\
 & 35 & 161.11 & 0.0710 & 0.0541 \\
 \hline
 & avg & 126.77 & 0.0568 & 0.0433 \\
\hline
\multirow{7}{*}{\shortstack{SEMF\\(Ours)}} 
 & 1  & 107.24 & 0.0357 & 0.0347 \\
 & 3  & 106.32 & 0.0360 & 0.0351 \\
 & 7  & 103.38 & \textcolor{red}{\textbf{0.0353}} & \textcolor{red}{\textbf{0.0345}}\\
 & 14 & \textcolor{red}{\textbf{100.50}} & \textcolor{red}{\textbf{0.0342}} & \textcolor{red}{\textbf{0.0338}} \\
 & 21 & \textcolor{red}{\textbf{107.52}} & \textcolor{red}{\textbf{0.0369}} & \textcolor{red}{\textbf{0.0367}} \\
 & 35 & \textcolor{red}{\textbf{117.34}} & \textcolor{red}{\textbf{0.0386}} & \textcolor{red}{\textbf{0.0399}} \\
 \hline
 & avg & 107.05 & \textcolor{red}{\textbf{0.0361}} & \textcolor{red}{\textbf{0.0358}}\\
\hline
\end{tabularx}
\caption{Horizon-wise forecasting performance for Gold price forecasts. Best results are shown in \textbf{\textcolor{red}{red}}.}
\label{tab:horizon_wise_results}
\end{table}

Table~\ref{tab:horizon_wise_results} presents horizon-wise forecasting results for the Gold price prediction task.
Baseline time series models exhibit increasing errors as the forecasting horizon extends, which reflects limited long-range stability.
By comparison, SEMF shows smaller performance variations as forecast length increases.
Accuracy remains relatively stable even at longer horizons, whereas competing methods display monotonic error growth.
Such differences in horizon sensitivity affect how forecasting models behave under multi-horizon planning requirements.

Beyond point accuracy, robustness across forecasting horizons carries direct economic implications.
In commodity markets, horizon-dependent instability can result in inconsistent inventory allocation, inefficient hedging adjustments, and increased operational uncertainty.
Stable forecasting performance across forecast lengths supports coherent planning strategies under uncertainty.
This allows a single forecasting framework to be applied across short- and long-term horizons without frequent recalibration.
From an operational perspective, such consistency reduces exposure to horizon-driven decision volatility.

\subsection{Analysis: Key Design Choices}
We present an ablation analysis that examines the contribution of individual components within the proposed SEMF framework, with evaluation restricted to Gold price forecasting to ensure a controlled and interpretable setting.
Gold is selected as the representative asset due to its high liquidity, stable data availability, and well-characterized market behavior, which allows for clear identification of architectural effects under consistent data conditions.

\subsubsection{Impact of Image Transformation}
Table~\ref{tab:image_ablation} summarizes the forecasting performance of SEMF under four image transformations: line plot, Short-Time Fourier Transform (STFT), Complex Morlet wavelet (CMOR), and Morlet wavelet.
This comparison examines how shape-based and time-frequency representations affect forecasting accuracy under non-stationary dynamics.

Shape-based line plot representations produce the largest errors, reflecting the absence of explicit frequency information.
Time-frequency approaches yield improved performance, with fixed-window STFT providing moderate gains and adaptive wavelet-based representations offering further improvements.
Among the evaluated methods, the Morlet wavelet achieves the most accurate and stable forecasts across all metrics.
Performance differences across transformations remain consistent across evaluation measures, which underscores the importance of representation choice in spectrogram-based forecasting.
Wavelet-based representations offer greater flexibility in characterizing non-stationary temporal patterns than shape-based or fixed-window frequency encodings.

\begin{table}[!t]
\centering
\begin{tabularx}{\linewidth}{
    >{\centering\arraybackslash}X
    >{\centering\arraybackslash}X
    >{\centering\arraybackslash}X
    >{\centering\arraybackslash}X
}
\toprule
\textbf{Image} & \textbf{RMSE} & \textbf{RMAE} & \textbf{MAPE} \\
\midrule
Line   & 187.98 & 0.0634 & 0.0617 \\
STFT   & 173.20 & 0.0554 & 0.0544 \\
CMOR   & 129.93 & 0.0495 & 0.0495 \\
Morlet & 107.05 & 0.0361 & 0.0358 \\
\bottomrule
\end{tabularx}
\caption{Forecasting performance of SEMF with different time-frequency representations.}
\label{tab:image_ablation}
\end{table}

\subsubsection{Role of Exogenous Feature Encoding}
Table~\ref{tab:raw_encoder_ablation} compares forecasting performance under different encoders for exogenous variables.
A simple multilayer perceptron exhibits limited effectiveness in modeling these signals, indicating insufficient capacity to capture temporal dependencies and interactions across variables. 
In contrast, the Transformer-based encoder learns more informative representations, as its self-attention mechanism enables structured modeling of temporal patterns and variable-wise dependencies. 
This richer representation is more effectively integrated through the cross-attention fusion module, where exogenous features interact with spectral representations of the target series. 
As shown in Table~\ref{tab:raw_encoder_ablation}, substantial improvements in forecasting performance are observed compared to the MLP-based alternative.

\begin{table}[!t]
\centering
\begin{tabularx}{\linewidth}{
    >{\centering\arraybackslash}X
    >{\centering\arraybackslash}X
    >{\centering\arraybackslash}X
    >{\centering\arraybackslash}X
}
\toprule
\textbf{} & \textbf{RMSE} & \textbf{RMAE} & \textbf{MAPE}\\
\midrule
MLP & 273.17 & 0.1097 & 0.1095 \\
Transformer & 107.05 & 0.0361 & 0.0358 \\
\bottomrule
\end{tabularx}
\caption{Comparison of SEMF performance using different Exogenous Feature Encoders (MLP vs. Transformer).}
\label{tab:raw_encoder_ablation}
\end{table}

\subsubsection{Effect of Cross-Modal Alignment}
Cross-modal alignment strategies differ substantially in how effectively they integrate spectral and temporal information.
Table~\ref{tab:ca_ablation} compares their impact on forecasting performance in SEMF.
A single-direction cross-attention (CA) mechanism allows only limited interaction between spectral and temporal representations, which constrains the effective use of complementary information across modalities.
In contrast, bidirectional cross-attention (bi-CA) enables mutual information exchange between the two representations.
This bidirectional alignment yields more informative fused representations and supports stable forecasting behavior.
Such alignment mechanisms enable complementary spectral and temporal signals to be utilized more effectively within a unified representation for multimodal forecasting.

\begin{table}[!t]
\centering
{\setlength{\tabcolsep}{6pt} 
\begin{tabularx}{\linewidth}{
    >{\centering\arraybackslash}X
    >{\centering\arraybackslash}X
    >{\centering\arraybackslash}X
    >{\centering\arraybackslash}X
}
\toprule
\textbf{Fusion} & \textbf{RMSE} & \textbf{RMAE} & \textbf{MAPE} \\
\midrule
Single CA & 200.66 & 0.0766 & 0.0776 \\
Bi-CA & 107.05 & 0.0361 & 0.0358  \\
\bottomrule
\end{tabularx}}
\caption{Forecasting performance of SEMF with different cross-attention fusion strategies.}
\label{tab:ca_ablation}
\end{table}

\begin{table}[!t]
\centering
\begin{tabularx}{\linewidth}{
    >{\centering\arraybackslash}X
    >{\centering\arraybackslash}X
    >{\centering\arraybackslash}X
    >{\centering\arraybackslash}X
    >{\centering\arraybackslash}X
}
\toprule
\textbf{Patch} & \textbf{Scale} & \textbf{RMSE} & \textbf{RMAE} & \textbf{MAPE} \\
\midrule
8  & 64  & 199.32 & 0.0696 & 0.0670 \\
16 & 64  & 142.79 & 0.0536 & 0.0524 \\
8  & 128 & 107.05 & 0.0361 & 0.0358 \\
16 & 128 & 181.61 & 0.0615 & 0.0587 \\
\bottomrule
\end{tabularx}
\caption{Forecasting performance of SEMF under varying patch sizes and Morlet wavelet scales.}
\label{tab:patch_scale_ablation}
\end{table}

\subsubsection{Design Trade-offs in Temporal-Spectral Resolution}
Table~\ref{tab:patch_scale_ablation} reports the forecasting performance of SEMF under different combinations of patch size and Morlet wavelet scale.
These parameters jointly determine the encoding of temporal and spectral information within the spectrogram representation and influence forecasting performance.


Patch size controls the granularity of temporal segmentation. 
Smaller patches preserve fine-grained temporal variations, whereas larger patches aggregate information over longer intervals and reduce temporal resolution. 
Wavelet scale governs the extent of spectral context. 
Larger scales capture longer-term frequency patterns, while smaller scales focus on short-term spectral components.

The results suggest a trade-off between temporal resolution and spectral context. 
The combination of a small patch size and a large wavelet scale yields the best forecasting performance, as it preserves localized temporal variations while providing sufficient long-range spectral information. 
In contrast, configurations with both large patch sizes and large wavelet scales result in overly coarse representations, which are associated with degraded performance due to the loss of fine-grained temporal structure. 
For smaller wavelet scales, the limited spectral context constrains the capacity of the learned representation to capture long-term frequency dynamics, even with preserved temporal resolution.

\begin{table}[!t]
\centering
\begin{tabularx}{\linewidth}{
    >{\centering\arraybackslash}X
    >{\centering\arraybackslash}X
    >{\centering\arraybackslash}X
    >{\centering\arraybackslash}X
}
\toprule
\textbf{\mbox{Seq Length}} & \textbf{RMSE} & \textbf{RMAE} & \textbf{MAPE} \\
\midrule
30  & 269.94 & 0.1092 & 0.1101 \\
60  & 270.05 & 0.1090 & 0.1094 \\
90  & 244.43 & 0.0848 & 0.0821 \\
120 & 107.05 & 0.0361 & 0.0358 \\
\bottomrule
\end{tabularx}
\caption{Forecasting performance with different input sequence lengths.}
\label{tab:seq_length_ablation}
\end{table}

\subsubsection{Effect of Historical Window Size}
Table~\ref{tab:seq_length_ablation} summarizes the forecasting performance of SEMF under different input sequence lengths.
The historical window size determines the amount of temporal context available to the model and directly affects its ability to capture both short-term dependencies and long-term trends.
Shorter input sequences provide insufficient historical information and result in substantially degraded forecasting performance.
As the sequence length increases, forecasting accuracy improves, indicating that additional historical context contributes to more informative temporal representations.
The longest sequence length considered in this study provides sufficient context to support stable forecasting behavior across horizons.
These observations place practical constraints on the minimum historical coverage required for reliable long-range forecasting.

\section{Conclusion}
This study presents the \textit{Spectrogram-Enhanced Multimodal Fusion} (SEMF) framework for multivariate time series forecasting in complex, non-stationary environments.
SEMF integrates Morlet wavelet spectrograms of the target series with sequential representations of exogenous variables, which are processed through separate encoding pathways and integrated via bidirectional cross-attention.
This design enables the model to jointly capture time-frequency characteristics and multivariate temporal dependencies within a unified forecasting framework.
Empirical evaluation demonstrates that SEMF achieves consistent forecasting performance across multiple horizons and commodities.
Component-wise analyses further confirm the contribution of each design element by demonstrating the importance of spatial and temporal resolution choices such as patch size, wavelet scale, and historical window length.

Beyond predictive accuracy, the horizon-consistent behavior of SEMF carries meaningful implications for operational and managerial decision-making.
Stable performance across forecast horizons reduces sensitivity to planning horizon selection in applications such as procurement, inventory management, and risk mitigation under volatile market conditions.
Such consistency supports more reliable coordination between short-term operational actions and longer-term strategic planning.
Forecasting systems with unstable long-horizon behavior can introduce unnecessary replanning and increased decision volatility.
In this sense, SEMF offers coherent decision-oriented forecasting across multiple time horizons in business and financial domains.

The applicability of SEMF extends beyond the evaluated commodities to other asset classes and industrial settings characterized by complex temporal dynamics and heterogeneous external influences.
The framework is particularly relevant in contexts where forecasting models serve as inputs to downstream decision processes.
Future work will explore extensions to higher-frequency and streaming data, alternative spectral representations, and additional information sources such as textual or sentiment-based signals.
Further advances in the interpretability of multimodal attention mechanisms may also facilitate broader adoption in decision-critical applications.

\section{Acknowledgments}
This research was supported by the Ministry of Science and ICT (MSIT), Korea, under the National Program for Excellence in SW (2023-0-00055), supervised by the Institute for Information \& Communications Technology Planning \& Evaluation (IITP).

\bibliography{sypark-AAAI-2025}

@article{hochreiter1997long,
  title={Long short-term memory},
  author={Hochreiter, Sepp and Schmidhuber, J{\"u}rgen},
  journal={Neural computation},
  volume={9},
  number={8},
  pages={1735--1780},
  year={1997},
  publisher={MIT press}
}

@article{SEZER2020106181,
title = {Financial time series forecasting with deep learning : A systematic literature review: 2005–2019},
journal = {Applied Soft Computing},
volume = {90},
pages = {106181},
year = {2020},
issn = {1568-4946},
doi = {https://doi.org/10.1016/j.asoc.2020.106181},
url = {https://www.sciencedirect.com/science/article/pii/S1568494620301216},
author = {Omer Berat Sezer and Mehmet Ugur Gudelek and Ahmet Murat Ozbayoglu}
}

@article{semenoglou2023image,
  title={Image-based time series forecasting: A deep convolutional neural network approach},
  author={Semenoglou, Artemios-Anargyros and Spiliotis, Evangelos and Assimakopoulos, Vassilios},
  journal={Neural Networks},
  volume={157},
  pages={39--53},
  year={2023},
  publisher={Elsevier}
}

@article{chen2024visionts,
  title={Visionts: Visual masked autoencoders are free-lunch zero-shot time series forecasters},
  author={Chen, Mouxiang and Shen, Lefei and Li, Zhuo and Wang, Xiaoyun Joy and Sun, Jianling and Liu, Chenghao},
  journal={arXiv preprint arXiv:2408.17253},
  year={2024}
}

@article{torrence1998practical,
  title={A practical guide to wavelet analysis},
  author={Torrence, Christopher and Compo, Gilbert P},
  journal={Bulletin of the American Meteorological society},
  volume={79},
  number={1},
  pages={61--78},
  year={1998},
  publisher={American Meteorological Society}
}

@inproceedings{zeng2023pixels,
  title={From pixels to predictions: Spectrogram and vision transformer for better time series forecasting},
  author={Zeng, Zhen and Kaur, Rachneet and Siddagangappa, Suchetha and Balch, Tucker and Veloso, Manuela},
  booktitle={Proceedings of the Fourth ACM International Conference on AI in Finance},
  pages={82--90},
  year={2023}
}

@article{dosovitskiy2020image,
  title={An image is worth 16x16 words: Transformers for image recognition at scale},
  author={Dosovitskiy, Alexey},
  journal={arXiv preprint arXiv:2010.11929},
  year={2020}
}

@inproceedings{kim2021reversible,
  title={Reversible instance normalization for accurate time-series forecasting against distribution shift},
  author={Kim, Taesung and Kim, Jinhee and Tae, Yunwon and Park, Cheonbok and Choi, Jang-Ho and Choo, Jaegul},
  booktitle={International conference on learning representations},
  year={2021}
}

@article{taylor2018forecasting,
  title={Forecasting at scale},
  author={Taylor, Sean J and Letham, Benjamin},
  journal={The American Statistician},
  volume={72},
  number={1},
  pages={37--45},
  year={2018},
  publisher={Taylor \& Francis}
}

@article{liu2023itransformer,
  title={itransformer: Inverted transformers are effective for time series forecasting},
  author={Liu, Yong and Hu, Tengge and Zhang, Haoran and Wu, Haixu and Wang, Shiyu and Ma, Lintao and Long, Mingsheng},
  journal={arXiv preprint arXiv:2310.06625},
  year={2023}
}

@article{nie2022time,
  title={A Time Series is Worth 64Words: Long-term Forecasting with Transformers},
  author={Nie, Y},
  journal={arXiv preprint arXiv:2211.14730},
  year={2022}
}

@article{wu2022timesnet,
  title={Timesnet: Temporal 2d-variation modeling for general time series analysis},
  author={Wu, Haixu and Hu, Tengge and Liu, Yong and Zhou, Hang and Wang, Jianmin and Long, Mingsheng},
  journal={arXiv preprint arXiv:2210.02186},
  year={2022}
}

@article{vaswani2017attention,
  title={Attention is all you need},
  author={Vaswani, Ashish and Shazeer, Noam and Parmar, Niki and Uszkoreit, Jakob and Jones, Llion and Gomez, Aidan N and Kaiser, {\L}ukasz and Polosukhin, Illia},
  journal={Advances in neural information processing systems},
  volume={30},
  year={2017}
}

@inproceedings{zhou2021informer,
  title={Informer: Beyond efficient transformer for long sequence time-series forecasting},
  author={Zhou, Haoyi and Zhang, Shanghang and Peng, Jieqi and Zhang, Shuai and Li, Jianxin and Xiong, Hui and Zhang, Wancai},
  booktitle={Proceedings of the AAAI conference on artificial intelligence},
  volume={35},
  number={12},
  pages={11106--11115},
  year={2021}
}

@article{wu2021autoformer,
  title={Autoformer: Decomposition transformers with auto-correlation for long-term series forecasting},
  author={Wu, Haixu and Xu, Jiehui and Wang, Jianmin and Long, Mingsheng},
  journal={Advances in neural information processing systems},
  volume={34},
  pages={22419--22430},
  year={2021}
}

@inproceedings{tsai2019multimodal,
  title={Multimodal transformer for unaligned multimodal language sequences},
  author={Tsai, Yao-Hung Hubert and Bai, Shaojie and Liang, Paul Pu and Kolter, J Zico and Morency, Louis-Philippe and Salakhutdinov, Ruslan},
  booktitle={Proceedings of the conference. Association for computational linguistics. Meeting},
  volume={2019},
  pages={6558},
  year={2019}
}

@inproceedings{zhang2023crossformer,
  title={Crossformer: Transformer utilizing cross-dimension dependency for multivariate time series forecasting},
  author={Zhang, Yunhao and Yan, Junchi},
  booktitle={The eleventh international conference on learning representations},
  year={2023}
}

@article{li2023time,
  title={Time series as images: Vision transformer for irregularly sampled time series},
  author={Li, Zekun and Li, Shiyang and Yan, Xifeng},
  journal={Advances in Neural Information Processing Systems},
  volume={36},
  pages={49187--49204},
  year={2023}
}

\end{document}